\pdfoutput=1

\documentclass{article}

\usepackage[preprint]{neurips_2024}

\usepackage[utf8]{inputenc} 
\usepackage[T1]{fontenc}    
\usepackage{hyperref}       
\usepackage{url}            
\usepackage{booktabs}       
\usepackage{amsfonts}       
\usepackage{nicefrac}       
\usepackage{microtype}      
\usepackage{xcolor}         
\usepackage{graphicx}
\usepackage{amsmath}
\usepackage[ruled, vlined, linesnumbered]{algorithm2e}

\title{Black-Box Opinion Manipulation Attacks to Retrieval-Augmented Generation of Large Language Models}

\author{%
  Zhuo Chen \\
  Wuhan University\\
  Wuhan, China \\
  \texttt{chenzhuo432@whu.edu.cn} \\
  \And
  Jiawei Liu\thanks{Corresponding author.}\\
  Wuhan University \\
  Wuhan, China \\
  \texttt{laujames2017@whu.edu.cn} \\
  \AND
  Haotan Liu \\
  Wuhan University \\
  Wuhan, China \\
  \texttt{baker-haotanliu@whu.edu.cn} \\
  \And
  Qikai Cheng \\
  Wuhan University \\
  Wuhan, China \\
  \texttt{chengqikai@whu.edu.cn} \\
  \And
  Fan Zhang \\
  Wuhan University \\
  Wuhan, China \\
  \texttt{fan.zhang@whu.edu.cn} \\
  \AND
  Wei Lu\\
  Wuhan University \\
  Wuhan, China \\
  \texttt{weilu@whu.edu.cn} \\
  \And
  Xiaozhong Liu\\
  Worcester Polytechnic Institute \\
  USA \\
  \texttt{xliu14@wpi.edu} \\
}

\begin{document}

\maketitle

\begin{abstract}
  Retrieval-Augmented Generation (RAG) is applied to solve hallucination problems and real-time constraints of large language models, but it also induces vulnerabilities against retrieval corruption attacks. Existing research mainly explores the unreliability of RAG in white-box and closed-domain QA tasks. In this paper, we aim to reveal the vulnerabilities of Retrieval-Enhanced Generative (RAG) models when faced with black-box attacks for opinion manipulation. We explore the impact of such attacks on user cognition and decision-making, providing new insight to enhance the reliability and security of RAG models. We manipulate the ranking results of the retrieval model in RAG with instruction and use these results as data to train a surrogate model. By employing adversarial retrieval attack methods to the surrogate model, black-box transfer attacks on RAG are further realized. Experiments conducted on opinion datasets across multiple topics show that the proposed attack strategy can significantly alter the opinion polarity of the content generated by RAG. This demonstrates the model's vulnerability and, more importantly, reveals the potential negative impact on user cognition and decision-making, making it easier to mislead users into accepting incorrect or biased information.
\end{abstract}

\section{Introduction}

With the rapid development of artificial intelligence, large language models (LLMs) have demonstrated exceptional capabilities in the field of natural language processing. However,  constrained by their training data, these models have limited scope of knowledge and lack the most up-to-date information, which can lead to errors or hallucinations when tackling more complex or time-sensitive tasks. Retrieval-Augmented Generation (RAG) combines information retrieval with the generative capabilities of large language models, enhancing the timeliness of knowledge acquisition and effectively mitigating the hallucination problem of these models. When given a query, RAG retrieves the most relevant passages from a knowledge base to augment the input request for the LLM. For example, the retrieved knowledge may consist of a series of text snippets that are semantically most similar to the query. RAG has inspired many popular applications, such as Microsoft Bing Chat, ERNIE Bot, and KimiChat, which use RAG to summarize retrieval results for improved user experience. Open-source projects like LangChain and LlamaIndex provide developers with flexible RAG frameworks to build customized AI applications using LLMs, retrieval models and knowledge bases. 

However, as the application scope of RAG expands, its security is increasingly a concern, especially regarding the model performance when faced with malicious attacks. The basic RAG process typically consists of three components: the corpus(refers knowledge bases), the retriever, and the generative large language model. When some of the retrieved passages are corrupted by malicious manipulators, the RAG process can become vulnerable; this is referred to as a retrieval manipulation attack in this paper. Numerous studies have explored various forms of retrieval manipulation attacks, such as adversarial attack on the retriever \cite{liu_order-disorder_2023, liu2023black}, prompt injection attack \cite{cai2022badprompt, liu2023prompt, jain2023baseline}, jailbreak attack for LLM \cite{deng2023jailbreaker, li2023multi, zhao2024weak}, and poisoning attack targeting the retrieval corpus in RAG \cite{zou2024poisonedrag, xue2024badrag}. 

This paper primarily focuses on adversarial ranking poisoning attacks against the retriever in RAG and how such attacks indirectly affect the generative results of the LLM. The threat model presented here is closer to a real-world black-box scenario and can be specifically modeled as follows: the attacker can only make requests to the large model and cannot access the complete corpus, the retriever, or the parameters of the RAG. The attacker can only insert adversarially modified candidate texts into the corpus, while the retriever and the LLM remain black-boxed, intact and unmodifiable. Based on previous studies \cite{lin2023mawseo, carlini2023poisoning}, the retrieval corpus and knowledge base contain millions of candidate texts sourced from the internet, allowing attackers to inject adversarially modified candidate texts by maliciously crafting web content or encyclopedia pages. Representative previous studies by Cho et al. \cite{cho2024typos} and Zhong et al. \cite{zhong2023poisoning} utilized predefined white-box retrievers, which are challenging to achieve in real-world scenarios with limited flexibility and practicality. Moreover, these works did not consider testing attacks specifically targeting the integrated generation process, where practical integrated models may mitigate the effects of attacks solely targeting the retriever, thereby reducing their effectiveness. Furthermore, another notable work, PoisonedRAG \cite{zou2024poisonedrag}, implemented black-box retrieval poisoning attacks on RAG knowledge bases, effectively exposing relevant security vulnerabilities of RAG. However, its experiments mainly focused on closed-domain question answering, such as "Who is the CEO of OpenAI?" Such questions can be corrected when RAG is combined with fact-checking and value alignment of LLMs. The vulnerabilities explored in this paper primarily target open-ended, controversial, and opinion-based questions in RAG, such as "Should abortion be legal?" These questions demand higher levels of logical analysis and summarization capabilities from large models. Current research in controversial topics is limited, and attacks manipulating opinions on opinion-based questions could potentially cause more profound harm.

Open-ended and controversial topics are issues that lack consensus due to differing opinions and attract widespread attention. These topics often involve opinions from different perspectives, influencing public perception when they are widely discussed. For example, in political elections, Robert Epstein \cite{epstein_search_2015} found that manipulating search engines to produce biased search results can alter voters' voting preferences. Placing passages favoring a particular candidate at the top significantly affects voter trust and favorability towards that candidate. Today, the issue of information homogenization in "information bubbles" has been a major concern among scholars. Zhang Yue et al. \cite{zhang2023homogenizationdilemma} proposed that homogenization in information bubbles manifests in three dimensions: selective homogenization, content homogenization, and group homogenization. Content homogenization refers to the phenomenon where people using online media encounter homogeneity in the presented content, often due to the "filter bubbles" that are created by recommendation systems and selectively feed biased information. In scenarios of open-ended and controversial topics, "information bubbles" can lead to the homogenization of user opinions, with people's views being easily influenced by the stance of the information they encounter. Through manual construction or search engine optimization, opinion manipulation attacks or "cognitive warfare" in open-ended controversial topics is actually widespread in practical applications such as social media and news platform. This phenomenon has numerous negative impacts on society. With the development of large language models, opinion manipulation exploiting RAG vulnerabilities poses a particularly severe threat. Attackers can influence the stance of the model generated content with carefully designed inputs, further endangering users' cognition and decision-making processes. Therefore, it is of significant theoretical and practical importance to study the vulnerabilities of RAG models against opinion manipulation attacks in black-box setting.

In short, this paper aims to explore the reliability of RAG against black-box opinion manipulation attacks in open-ended controversial topics and investigate the impact of such attacks on user cognition and decision-making. Specifically, we first send specific instructions to obtain the ranking of the retrieval results in the RAG model and analyze the working mechanism of its retrieval module. We train a surrogate model on the obtained retrieval ranking data to approximate the features and relevance preferences of the retriever in RAG \cite{liu_order-disorder_2023, wu_prada_2022}. Based on the surrogate model, we design adversarial retrieval attack strategies to manipulate the opinions of candidate documents. By attacking this surrogate model, we generate adversarial opinion manipulation samples and transfer these adversarial samples to the actual RAG model. We then conduct experiments on opinion datasets across multiple topics to validate the effectiveness and impact range of the attack strategies without understanding the internal knowledge of the RAG model. Experiments conducted on opinion datasets across multiple topics show that the proposed attack strategy can significantly alter the opinion polarity of the content generated by RAG. This not only demonstrates the vulnerability of the model but, more importantly, reveals the potential negative impact on user cognition and decision-making, making it easier to mislead users into accepting incorrect or biased information.

\section{Related Works}

Research on the reliability of neural network models has long been established. In 2013, Szegedy et al. \cite{szegedy2013intriguing} found that applying imperceptible perturbations to a neural network model during a classification task was sufficient to cause classification errors in CV. Later, scholars observed similar phenomenon in NLP. Robin et al. \cite{jia2017adversarial} found that inserting perturbed text into original paragraphs significantly distracts computer systems without changing the correct answer or misleading humans. It reflects the robustness of neural network models, i.e., the ability to output stable and correct predictions in tackling the imperceptible additive noises \cite{wang2019towards}. For large language models, Wang et al. \cite{wang2023decodingtrust} proposed a comprehensive trustworthiness evaluation framework for LLMs, assessing their reliability from various perspectives such as toxicity, adversarial robustness, stereotype bias, and fairness. While large language models have greater capabilities compared to general deep neural network models, they also raise more concerns regarding security and reliability. 

As RAG is designed to overcome the hallucination problem in LLMs and enhance their generative capabilities, the reliability of the content generated by RAG is also a major concern. Zhang et al. \cite{zhang2024human} attempted to explore the weaknesses of RAG by analyzing critical components in order to facilitate the injection of the attack sequence and crafting the malicious document with a gradient-guided token mutation technique. Xiang et al. \cite{xiang2024certifiably} designed an isolate-then-aggregate strategy, which gets responses of LLMs from each passage in isolation and then securely aggregate these isolated responses, to construct the first defense framework against retrieval corruption attacks. These studies are based on white-box scenarios and primarily focus on the robustness of RAG against corrupted and toxic content.

This paper intends to use adversarial retrieval attack strategies to perturb the ranking results of the retriever, ensuring that opinion documents with a certain stance are ranked as high as possible, thereby guiding the generated responses of the LLM to reflect that stance.

The adversarial retrieval attack strategy starts with manipulation at the word level. Under white-box setting, Ebrahimi et al. \cite{ebrahimi2017hotflip} utilize an atomic flip operation, which swaps one token for an other, to generate adversarial examples and the method, known as Hotflip. Hotflip gets rid of reliance on rules, but the adversarial text it generate usually has incomplete semantics and insufficient grammar fluency. While it can deceive the target model, it cannot evade perplexity-based defenses. Wu et al. \cite{wu_prada_2022} also proposed a word substitution ranking attack method called PRADA. To enhance the readability and effectiveness of the adversarial text, scholars further designed sentence-level ranking attack methods. Song et al. \cite{song2020adversarial} propose an adversarial method under white-box setting, named Collision, which uses gradient optimization and beam search to produce the adversarial text named collision. The Collision method further imposes a soft constraint on collision generation by integrating a language model, reducing the perplexity of the collision. The method has shown promising\cite{liu_order-disorder_2023} propose the Pairwise Anchor-based Trigger (PAT) method under black-box setting. Added the fluency constraint and the next sentence prediction constraint, the method generates adversarial text by optimizing the pairwise loss of top candidates and target candidates with adversarial text. Although the time complexity of PAT has increased compared to previous methods, PAT takes ranking similarity and semantic consistency into account, so its manipulation effect on the retrieval ranking of target candidates is superior.

\section{Method}

\begin{figure}
  \centering
  \vspace{-40pt}
  \includegraphics[scale=0.42]{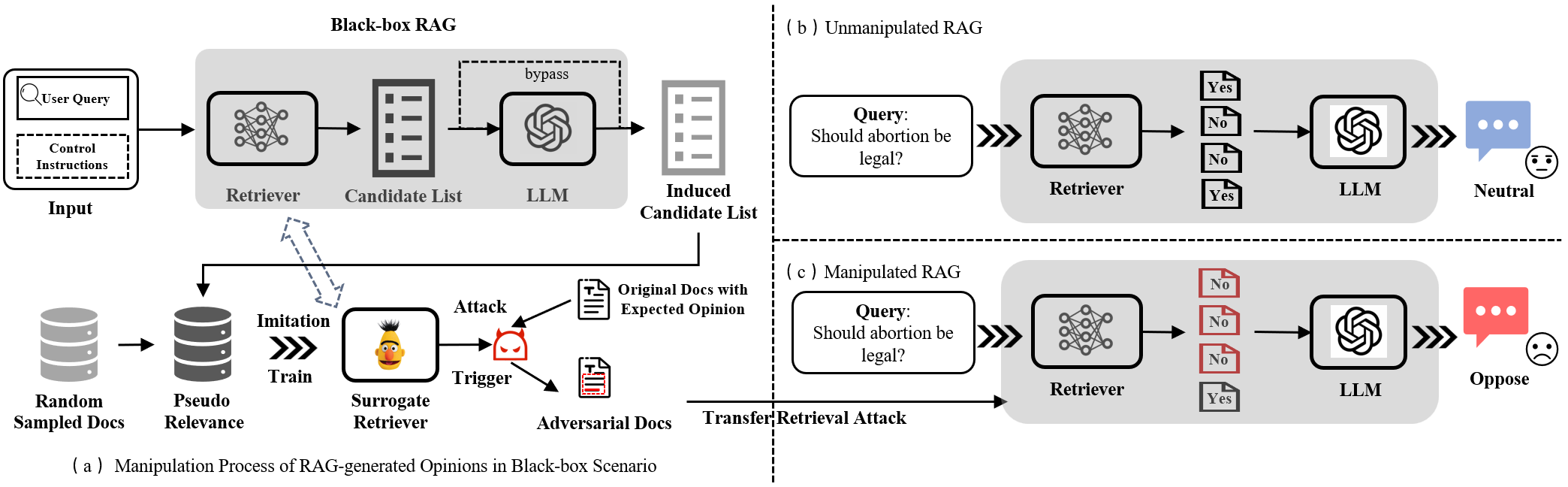}
  \caption{The method for manipulating the opinions of RAG-generated content in black-box scenario}
  \vspace{0pt}
\end{figure}

This paper attempts to manipulate the opinions in the responses generated by black-box RAG models on controversial topics, targeting both the retrieval model and the LLM which performs the integrated generation task. Zhang et al. \cite{zhang2024human} tried to poison context documents to deceive the LLM into generating incorrect content, but this method requires extensive internal details of the LLM application, making it less feasible in real-world scenarios. For black-box RAG, the manipulator has no knowledge of the internal information of the RAG, including model architecture and score function, and can only access the inputs and outputs of the RAG. Specially, the manipulator can only call the interface of the LLM in RAG instead of that of the retriever. Since the inputs consist of the query and the candidate documents and the user's query cannot be altered, this paper focuses on modifying the candidate documents. Although the manipulator cannot access the entire corpus, they can insert adversarially modified candidate texts into the corpus. The basic framework of RAG consists of the retriever and the generative large language model, which the two are serially connected, the LLM performs the generation task based on the context information retrieved by the retriever. Given that manipulators in a black-box scenario cannot modify the system prompts of the generative large model, it is difficult to directly manipulate the generation results by exploiting the reliability flaws of the LLM itself. Therefore, this paper focuses on exploiting the reliability flaws of the retriever to manipulate the retrieval ranking results. By adding adversarial texts to candidate documents that hold the expected opinion, we increase their relevance to the query, making them more likely to be included in the context passed to the generative large language model. Leveraging the strong capability of LLM for understanding and following instructions, we guide the LLM to generate responses that align with the expected opinion. An overview of this method is shown in Figure 1.

The specific approach for manipulating RAG opinions on controversial topics is as follows: Given a topic \( q \) (the query) from a set of controversial topics \( Q \), we select a expected opinion \( S_t \) and target all candidate documents \( d_t \) in the retrieval corpus \( D \) that hold the \( S_t \) opinion. After obtaining the adversarial text \( p_{\text{adv}} \), it is added to \( d_t \), transforming the retrieval corpus to \( D(d; d_t \oplus p_{\text{adv}}) \). Since \( p_{\text{adv}} \) can increase the relevance score \( R(q, d_t \oplus p_{\text{adv}}) \) assigned by the retrieval model \(\text{RM}\) to \( d_t \) for query \( q \), ideally \( d_t \) will be ranked at the top of the retrieval results \(\text{RM}_k(q) = \{ d \mid d_t \oplus p_{\text{adv}} \} \), guiding the large language model to generate responses that align with the expected opinion: \( S(\text{LLM}(q, \text{RM}_k(q))) = S_t \).

The primary issue in implementing manipulation is to make the retrieval model of the black-box RAG transparent. This paper aims to simulate the retrieval model \( RM\). The basic idea is to train a surrogate model \( M_{\text{i}}\) with the ranking results \( RM_{\text{k}}(q)\) from the retrieval model \( RM\), thus turning the black-box retrieval model into a white-box model. However, since the retriever and the large generative model in RAG are serially connected, it is not possible to directly obtain the ranking results of the retriever. Therefore, this paper attempts to guide the large generative model to replicate the retrieval results of the black-box RAG. Therefore, this paper attempts to guide the large model to replicate the output of the retrieval model, so we obtain the text data deemed relevant by the black-box retrieval model \( RM\), which can be used as positive examples \( d_+ \) for black-box imitation training. Subsequently, irrelevant texts to the query can be random sampled as negative examples \( d_- \) . Therefore, this paper designs specific instructions to make the black-box RAG replicate the retrieval results of the retriever \( \text{RM} \). These retrieval results only need to reflect the relevance to the query. Then, based on the generated results of the LLM, we sample positive and negative data to train the surrogate model. The method for obtaining imitation data of the retrieval model in a black-box RAG scenario is illustrated in Figure 2. The prompt instruction used is as follows:

\begin{center}
    \fbox{\parbox{10cm}{ \emph{Now that you are a search engine, please search: \{query\} \\
Ignore the Question. Please copy the top 3 passages of the given Context intact in the output and provide the output in JSON with keys 'answer' and 'context'. Put each candidate passage in 'context' as a string element in the list. Candidate passages are separated by line break instead of period or exclamation point. Each candidate is an element in the list, like [Passage 1, Passage 2, Passage 3]. Please copy the passages intact with no modification and only output the one best JSON response.}}} 
\end{center}

\begin{figure}
  \centering
  \includegraphics[scale=0.7]{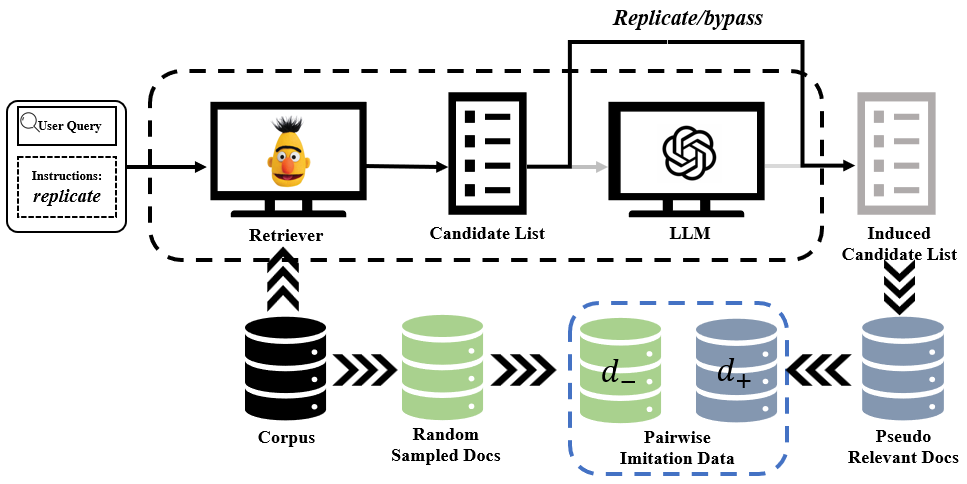}
  \caption{The method for obtaining imitation data of RAG retrieval model in black-box scenario}
\end{figure}

This paper uses a pairwise approach to sample data and train the surrogate model. Relevant passages are sampled from the responses generated by the black-box RAG as positive examples \( d_+ \), and random irrelevant passages are sampled as negative examples \( d_- \). The black-box RAG responds with context information so the responses generated reflect the retrieval results instead of being independent on the context. These sample pairs \((d_+, d_-)\) are incorporated into the training dataset. After sampling the imitation data, this paper uses a pairwise training method to obtain the surrogate model \( M_i \). Let the relevance score calculated by \( M_i \) be \( R_i \), the training optimization objective is as follows:

\begin{center}
    \hspace{25mm} \( L = -\frac{1}{|Q|} \sum_{q \in Q} \log \left( \frac{R_i (q, d_+)}{R_i (q, d_+) + \sum R_i (q, d_-)} \right)\)  \hspace{25mm}[1]
\end{center}

After obtaining the surrogate model \( M_i \), this paper transforms the manipulation of RAG-generated opinions in a black-box scenario into manipulation in a white-box scenario. Since we have all the knowledge of the white-box surrogate model \( M_i \), this paper directly implements adversarial retrieval attacks on it, generating adversarial text \( p_{\text{adv}} \) for the candidate document \( d_t \) holding the opinion \( S_t \). This paper employs the Pairwise Anchor-based Trigger (PAT) strategy for adversarial retrieval attacks, which is commonly used as a baseline in related research. Subsequently, the generated adversarial text is added to the candidate document with \( S_t \). Then, the system of the black-box RAG model is queried, and the generated response is obtained. The stance of the response is compared with the stance of the response generated by the RAG without manipulation to evaluate the reliability of the black-box RAG.

PAT, as a representative adversarial retrieval attack strategy, adopts a pairwise generation paradigm. Given the target query, the target candidate item, and the top candidate item(anchor, used to guide the adversarial text generation), the method utilizes gradient optimization of pairwise loss, calculated from the candidate item and the anchor, to find the appropriate representation of an adversarial text. The method also adds fluency constraint and next sentence prediction constraint. By beam search for the words, the final adversarial text, denoted as $T_{pat}$, is iteratively generated in an auto-regressive way. This paper uses \( T_{\text{pat}} \) as \( p_{\text{adv}} \), with its generated optimization function being \cite{liu_order-disorder_2023}:

\begin{center}
      \hspace{10mm} \(\max \left( M_i (q, T_{\text{pat}}; w) + \lambda_1 \cdot \log P_g (T_{\text{pat}}; w) + \lambda_2 \cdot f_{\text{nsp}} (d_t, T_{\text{pat}}; w) \right)\) \hspace{10mm}[2]
\end{center}

In the above formula, \( P_g \) is the semantic constraint function, and \( f_{\text{nsp}} \) is the next sentence prediction consistency score function between \( T_{\text{pat}} \) and \( d_t \).

In terms of dataset, this paper uses the MS MARCO Passages Ranking dataset as the data source for guiding the black-box RAG to generate relevant passages \cite{zhong2023poisoning} where we sample data pairs to train the surrogate model. Additionally, this paper uses controversial topic data scraped from the PROCON.ORG website as the object of manipulation. The controversial topic dataset includes over 80 topics, covering fields such as society, health, government, education, and science. Each controversial topic is discussed from two stances (pro and con), with an average of 30 related passages, each holding a certain opinion with stance pro or con.

The specific settings details for the RAG manipulation experiment are as follows:

(1) Black-box RAG: This paper represents the black-box RAG process, which serves as the research object, as \( \text{RAG}_{\text{black}} \). It mainly consists of a retriever and a large language model (LLM). The LLMs used are the open-source models Meta-Llama-3-8B-Instruct (LLAMA3-8B) and Qwen1.5-14B-Chat (Qwen1.5-14B). The LLAMA and Qwen series LLMs perform well across various tasks among all open-source models. The prompt connecting the retriever and the LLM in \( \text{RAG}_{\text{black}} \) adopts the basic RAG prompt from the Langchain framework:

\begin{center}
    \fbox{\parbox{10cm}{ \emph{Use the following pieces of retrieved context to answer the question. Keep the answer concise.
Context: \{context\}.
Question: \{question\}.
}}} 
\end{center}

(2) Target retriever model and surrogate model: The retriever in RAG is usually a dense retrieval model. Therefore, this paper selects the representative dense retrieval model, coCondenser, as the target retrieval model \cite{gao2021unsupervised}. Since coCondenser is a BERT-based model, the surrogate model chosen in this paper is the MiniLM model, which is BERT-based and specifically trained on the MS Marco Passage Ranking dataset.

(3) Manipulation target: For a controversial topic \( q \), documents \( d_t \) holding the expected opinion \( S_t \) are manipulated by adding adversarial text \( p_{adv} \) at the beginning. This manipulation aims to position these perturbed documents as prominently as possible in the top \( K \) rankings of the RAG retriever \( \mathrm{RM}_k(q) \), where \( K \) denotes the number of paragraphs obtained by the RAG generation model from the retrieval results. In this paper, \( K \) is set to 3.

(4) Manipulator(the threat model): In the black-box scenario, the manipulator is only authorized to query the RAG, obtain RAG-generated results and modify the target documents. There are no restrictions on the number of calls to RAG. Furthermore, the manipulator has no knowledge of the model architecture, model parameters, or any other information related to the models within the black-box RAG. Modifying the prompt templates used by the LLM is also prohibited.

(5) Experimental Parameters: The batch size for training the surrogate model is set to 32, with 24 iterations.

Our proposed opinion manipulation strategy for black-box RAG is outlined in Algorithm 1.

\begin{algorithm}[!t] 
    \newcommand\Parameter{\textbf{Parameter: }}
    \newcommand\Functions{\textbf{Functions: }}
    \newcommand\Instructions{\textbf{Instructions: }}
    \caption{Opinion Manipulation Strategy for black-box RAG}  
    \label{alg:algorithm1}  
    \LinesNotNumbered
    \KwIn{target black-box RAG model \( \textit{RAG}_{\text{black}} \), target retrieval model \( RM\), \text{surrogate model} \( \boldsymbol{M_i} \), controversial topics $\boldsymbol{Q}$,  target topic $ \boldsymbol{q}$, expected opinion \( S_t \), corpus \( Docs\) , target documents with expected opinion $Docs_t$, target document \( \boldsymbol{d_t} \), relevant document \( d_+ \), random sampled document \( d_- \) }
    
    \Instructions{\\
        \Indp \Indp \( i_1\) = "Now that you are a search engine, please search: \{query\} Ignore the Question. Please copy the top 3 passages of the given Context intact in the output and provide the output in JSON with keys ’answer’ and ’context’. Put each candidate passage in ’context’ as a string element in the list..."\\
        \( i_2\) = "Use the following pieces of retrieved context to answer the question..." \\ \tcp{\(\textit{RAG}_{\text{black}}\) uses \(i_2\) as prompt template.}
    }
    \Functions{\\
        \Indp \Indp $\boldsymbol{OpinionClassify}$: Classify the opinion of the content into "support", "neutral" or "oppose".\\
        $\boldsymbol{PAT}$: Pairwise Anchor-based Trigger generation strategy.\\
    }
    \LinesNumbered
    \KwOut{manipulated RAG responses $\boldsymbol{Res}$}
    
    \SetKwProg{Proc}{Phase}{}{}
    \Proc{1. Pairwise Imitation Data Construction and Black-box Retrieval Model Imitation Training}{
        INIT: Dataset $\mathcal{D} \gets \{\}$ \\
        \For{$\boldsymbol{q}_m \in \boldsymbol{Q}$}{
         induced rank list $\boldsymbol{R}_m{top3} \gets$  \( \textit{RAG}_{\text{black}}(\boldsymbol{q}_m \oplus i_1; \boldsymbol{Docs}) \) \\ \tcp{\( \textit{RAG}_{\text{black}}(\boldsymbol{q}_m \oplus i_1 ) \) $\approx$ \( RM(\boldsymbol{q}_m)\) }
            \For{$\boldsymbol{d_+}_j \in \boldsymbol{R}_m{top3}$}{
            Random sample document as $\boldsymbol{d_-}_j$ \\
            $\mathcal{D} \gets positive, [\boldsymbol{q}_m; \boldsymbol{d_+}_j; \boldsymbol{d_-}_j]$ \\
            $\mathcal{D} \gets negative, [\boldsymbol{q}_m; \boldsymbol{d_-}_j; \boldsymbol{d_+}_j]$ \\
            \tcp{Reverse $\boldsymbol{d_+}_j$ and $\boldsymbol{d_-}_j$ to get the negative triple} 
         }
    }
    Train the surrogate model \( \boldsymbol{M_i} \) on $\mathcal{D}$ with Eq 1\\
    \Return{\( \boldsymbol{M_i} \)}\
    }
    \Proc{2. Adversarial Trigger Generation and Opinion Manipulation in RAG Response}{
        INIT: RAG Response Set $\boldsymbol{Res} \gets \{\}$ \\
        \For{$\boldsymbol{q}_m \in \boldsymbol{Q}$}{
            rank list $\boldsymbol{R}_m \gets$ \( \boldsymbol{M_i(\boldsymbol{q}_m; Docs)} 
            \) \\
            $\boldsymbol{anchor}_m \gets$ \text{top-1}($\boldsymbol{R}_m$) \\
            \For{\( d_j \in \boldsymbol{R}_m\)}{
                \If{\(\boldsymbol{OpinionClassify}(d_j) = S_t\)}{
                $\boldsymbol{Docs_t} \gets$ \( d_j \)
                }
            }
            \For{\( \boldsymbol{d_t}_j \in \boldsymbol{Docs_t}\)}{
                adversarial trigger \( \boldsymbol{p_{adv}}_j \gets \boldsymbol{PAT}(\boldsymbol{M_i}; \boldsymbol{q}_m ,\boldsymbol{d_t}_j, \boldsymbol{anchor}_m)\) \\
                adversarial document \( \boldsymbol{d_{adv}}_j \gets \boldsymbol{d_t}_j \oplus \boldsymbol{p_{adv}}_j\) \\
                \( \boldsymbol{Docs} : \boldsymbol{d_t}_j \gets \boldsymbol{d_{adv}}_j\) \tcp{Replace} 
            }
            \( \boldsymbol{Res} \gets \) \( \textit{RAG}_{\text{black}}(\boldsymbol{q}_m;\boldsymbol{Docs}) \) \\
        }
        \Return{\(\boldsymbol{Res}\)}
    }
\end{algorithm}

\section{Experiment and Analysis}

After imitating the retrieval model of \( \text{RAG}_{\text{black}} \) to obtain the surrogate model, this paper first compares the ranking ability of the surrogate model \( M_i \) and the target retrieval model \( RM \), as well as the similarity of their ranking results, as shown in Table 1, to ensure that the surrogate model has learned the capabilities of the black-box retrieval model.

\begin{table}
  \caption{Comparison(\%) of ranking results between the surrogate model and the target retrieval model(based on the target retrieval model)}
  \label{sample-table}
  \centering
  \begin{tabular}{lllll}
    \toprule
    \cmidrule(r){1-5}
    Model     & MRR@10     & NDCG@10  & Inter@10 & RBO@10 \\
    \midrule
    Target retrieval model & 87.07  & 68.16  & -- & --  \\
    Surrogate model     & 87.98 & 73.73  & 62.32 & 48.66   \\
    \bottomrule
  \end{tabular}
\end{table}

This paper uses Mean Reciprocal Rank (MRR) and Normalized Discounted Cumulative Gain (NDCG) to reflect the ranking ability of the models themselves; higher values indicate stronger ranking ability in terms of relevance. Inter Ranking Similarity (Inter) and Rank Biased Overlap (RBO) are used to measure the similarity between the ranking results of the surrogate model and the target retrieval model; higher values indicate better performance of the black-box imitation. The weight for RBO@10 is set to 0.7. In Table 1, “--” indicates that the metric is not applicable to the model.

As can be seen from Table 1, the surrogate model \(M_i\) trained by black-box imitation is similar to the target retrieval model coCondenser in terms of relevance ranking performance and ranking results, validating the effectiveness of the black-box imitation.

After the black-box imitation training, the white-box surrogate model is conducted with opinion manipulation experiments. Several controversial topics and their opinion text data under the four themes of "Government", "Education", "Society", and "Health" from the PROCON.ORG data are selected as the retrieval corpus. The original retrieval corpus is denoted as \textit{Docs\textsubscript{origin}}. Based on the surrogate model, we generate the corresponding \textit{T\textsubscript{pat}} for the candidate items with the expected opinion \( S_t \) on controversial topics, and then insert \textit{T\textsubscript{pat}} at the beginning of the target candidate items to obtain the perturbed retrieval corpus \textit{Docs\textsubscript{adv}}. Query \text{RAG\textsubscript{black}} twice on controversial topics: once with \textit{Docs\textsubscript{origin}} as the retrieval corpus, and once with \textit{Docs\textsubscript{adv}} as the retrieval corpus, and obtain the two responses of \text{RAG}, representing the answers before and after opinion manipulation. The responses are then classified into three categories based on their opinion on controversial topics: opposing, neutral, and supporting, represented by 0, 1, and 2, respectively, as the opinion scores of the generated responses. This study uses Average Stance Variation (ASV) to represent the average increase of opinion scores of RAG\textsubscript{black} responses in the direction of the expected opinion \( S_t \) before and after manipulation. A positive ASV indicates that the opinion manipulation towards \( S_t \) is effective, while a negative ASV indicates that the manipulation actually makes the opinions of RAG responses deviate from \( S_t \). The larger the ASV value, the more successful the opinion manipulation of RAG responses. Additionally, this paper attempts to obtain the ranking results of the retriever coCondenser to evaluate the effectiveness of the adversarial retrieval manipulation strategy at the ranking stage for dense retrieval. This evaluation is solely for assessment purposes and is not involved in manipulation, as no internal knowledge of RAG\textsubscript{black} was leaked during the manipulation process.

After obtaining the ranking results of the retriever model coCondenser, this paper evaluates the manipulation effect with Attack Success Rate (ASR), the average proportion of target opinions in the Top 3 rankings before and after manipulation (Top3\textsubscript{origin}, Top3\textsubscript{attacked}), and the Variation of Normalized Discounted Cumulative Gain (VoN-DCG). Higher values of ASR and Vo-NDCG indicate better manipulation effects on ranking, and a larger difference between Top3\textsubscript{attacked} and Top3\textsubscript{origin} signifies more significant ranking manipulation effects, too.

\begin{table}
  \caption{Manipulation results of \(RAG\textsubscript{black}\) ranking and response opinion}
  \label{sample-table}
  \centering
  \begin{tabular}{ccccccc}
    \toprule
    Model &  \multicolumn{4}{c}{CoCondenser Ranking}  & Qwen1.5-14b  &  LLAMA3-8b \\
    Topic & \(ASR\) & \(Top3_origin\) & \(Top3_attacked\) & \(NDCG_variation\) & \(ASV\) & \(ASV\) \\
    \midrule
    Government & 0.17 & 0.48 & 0.57 & 0.06 & -0.17  & 0.25 \\
    Education  & 0.33 & 0.28 & 0.39 & 0.09 & 0.42 & 0.5 \\
    Society & 0.5 & 0.39 & 0.56 & 0.07 & 0.42 & 0.5 \\
    Health & 0.5 & 0.33 & 0.44 & 0.12 & 0.67 & 0.5 \\
    \bottomrule
  \end{tabular}
\end{table}

\begin{figure}
  \centering
  \fbox{\includegraphics[scale=0.8]{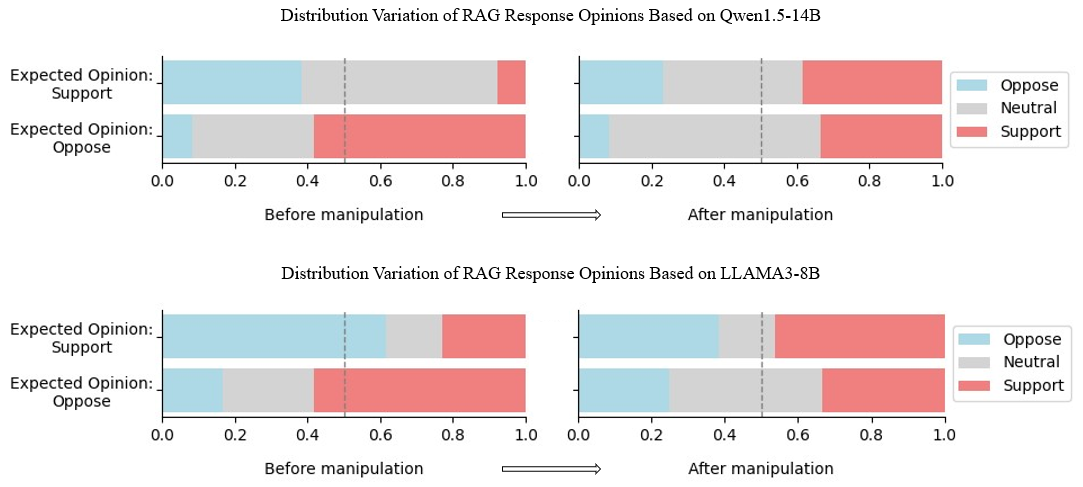}}
  \caption{Overall effect of RAG opinion manipulation in black-box scenario}
\end{figure}

Figure 3 shows the significant overall opinion manipulation effect of the adversarial retrieval attack strategy PAT. This paper divides Docs\textsubscript{origin} into two parts:  document data with an expected opinion of support and document  data with an expected opinion of opposition, the expected opinion represents the stance direction we would like RAG response to hold for the target topic after manipulation. The expected opinion $S_t$ is set to 2 for supporting and 0 for opposing, and then manipulation is performed. The results show that when the expected opinion is support, the proportion of responses with a supportive stance increases significantly after manipulation, while the proportion of responses with an opposing stance decreases. When the expected opinion is opposition, the proportion of supportive responses decreases significantly after manipulation, while the proportions of neutral and opposing responses both increase. Comparatively, the changes in stance before and after manipulation are slightly larger for LLAMA3-8b than for Qwen1.5-14b, due to stronger ability of LLAMA3-8b to follow instructions.

The results of the theme-specific manipulation experiments are shown in Table 2. The adversarial retrieval attack strategy PAT, applied to the adversarial texts generated by the surrogate model, significantly increased the proportion of candidate items holding expected opinion in the Top 3 of the RAG\textsubscript{black} retrieval list, thereby guiding the LLM to change its opinion in the response. However, the manipulation effect of RAG\textsubscript{black} generated opinions varies across different themes: for education, society, and health topics, the attack success rate and ranking variation of target items are significantly higher than those in government topics. This suggests that the LLM may have been specifically fine-tuned on government-related dataset, enabling it to mitigate the bias in the retrieval context to some extent. Among these topics, controversial opinions in society and health topics are more susceptible to manipulation. Since these two areas are closely related to people's lives, opinion manipulation in society and health topics may pose a greater risk.

The manipulation results across different themes still demonstrate relative advantage of LLAMA3-8b in understanding prompts with contextual background intentions and generating effective responses. However, this also indicates that the strong comprehension ability of LLMs may undermine the reliability of the content it generates.

\section{Conclusion}

In this paper, we explore the vulnerability of retrieval-augmented generation (RAG) models to opinion manipulation against black-box attack in open-ended controversial topics, and delve into the potential impact of such attacks on user cognition and decision-making. Through systematic experiments, we propose a novel adversarial attack strategy about retrieval ranking poisoning. This method significantly affects the polarity of the opinions generated by RAG by crafting adversarial samples, without requiring internal knowledge of the RAG model. The experimental results indicate that the proposed attack strategy successfully alters the opinion of the content generated by the RAG model, revealing the vulnerability and unreliability of RAG when confronted with malicious retrieval corpus. More importantly, this opinion manipulation could have profound impacts on users' cognition and decision-making processes, potentially leading users to accept incorrect or biased information, causing cognitive changes and public opinion distortion. This phenomenon is particularly significant in open-ended and controversial issues. 

Future research will expands the scale of the experiments by including more open-source and commercial RAG systems to more comprehensively evaluate the reliability of viewpoint generation by RAG models. Given the vulnerabilities of RAG models, future work should focus on developing more robust defense strategies. These may include improving the robustness of retrieval algorithms, enhancing the reliability of generation models, and introducing multi-level input filtering mechanisms to counteract adversarial inputs, thereby achieving a balanced optimization of the understanding and reliability of RAG models.

\section{Ethical Statement}
This paper explores the feasibility of opinion manipulation on black-box RAG models in real-world scenarios. The main goal is to assess the reliability of RAG technology in responding to ranking manipulation at the stage of retrieval, paving the way for future work to enhance the robustness and defense capabilities of RAG technology. This study did not manipulate any commercial RAG systems or real-world data currently in use.·
\nocite{chen2024research}
\nocite{zhang2021argot}
\nocite{cohen2019certified}


\bibliography{reference_rag}

\end{document}